\documentclass{article}

\usepackage[final]{neurips_2025}

\usepackage[utf8]{inputenc} 
\usepackage[T1]{fontenc}    
\usepackage{hyperref}       
\usepackage{url}            
\usepackage{booktabs}       
\usepackage{subcaption}

\usepackage{amsfonts}       
\usepackage{xspace}

\usepackage{nicefrac}       
\usepackage{microtype}      
\usepackage{xcolor}         
\usepackage{graphicx}
\usepackage{amsmath}
\usepackage{algorithm}      
\usepackage{algorithmicx}   
\usepackage{algpseudocode}  
\usepackage{multirow}
\usepackage{booktabs}
\usepackage{adjustbox}
\usepackage[table]{xcolor}
\usepackage{wrapfig}
\usepackage{enumitem}
\usepackage{colortbl}
\usepackage{xcolor}
\definecolor{darkblue}{rgb}{0, 0, 0.5}
\definecolor{darkgreen}{RGB}{50,100,0}
\definecolor{darkred}{RGB}{200, 0, 0}
\definecolor{lightblue}{RGB}{220,235,250}
\hypersetup{colorlinks=true, citecolor=darkblue, linkcolor=darkblue, urlcolor=darkblue}
\definecolor{lightblue}{RGB}{220,235,250}

\usepackage[most,skins,theorems]{tcolorbox}
\tcbset{
  takeawaysbox/.style={
    title=Takeaways,
    colback=lightblue!80,
    colframe=black,
    fonttitle=\bfseries\small,
    coltitle=white,
    colbacktitle=black,
    enhanced,
    attach boxed title to top left={xshift=2.5mm,yshift=-2.5mm},
    boxed title style={rounded corners, size=small, colframe=black, colback=black},
    width=\linewidth,
    arc=3.5mm
  }
}

\title{Advances and Innovations in the\\ Multi-Agent Robotic System (MARS) Challenge}

\author{%
   Li Kang$^{1,4\dagger}$, Heng Zhou$^{3,4\dagger}$, Xiufeng Song$^{1\dagger}$, Rui Li$^{4\dagger}$, Bruno N.Y. Chen$^{5}$, Ziye Wang$^{6}$,\\
   \textbf{Ximeng Meng$^{7}$, Stone Tao$^{8}$, Yiran Qin$^{9}$, Xiaohong Liu$^{1}$, Ruimao Zhang$^{10}$, Lei Bai$^{4}$,} \\
   \textbf{Yilun Du$^{11}$, Hao Su$^{8}$, Philip Torr$^{2}$, Zhenfei Yin$^{2\ddagger}$ and the Challenge Participants%
   \thanks{The complete list of participants is provided in Section~\ref{app:participants}.}} \\
  $^1$SJTU,
  ~$^2$Oxford,
  ~$^3$USTC,
  ~$^4$Shanghai AI Lab,
  ~$^5$CMU,
  ~$^6$HKU, \\
  $^7$Tongji, 
  ~$^8$UC San Diego,
  ~$^9$CUHK-SZ,
  ~$^{10}$SYSU,
  ~$^{11}$Harvard
  \\
  $^\dagger$ Equal contribution 
  $^\ddagger$ Corresponding author \\
}

\begin{document}

\maketitle

\begin{abstract}
Recent advancements in multimodal large language models and vision-language-action models have significantly driven progress in Embodied AI. 
As the field transitions toward more complex task scenarios, multi-agent system frameworks are becoming essential for achieving scalable, efficient, and collaborative solutions. This shift is fueled by three primary factors: increasing agent capabilities, enhancing system efficiency through task delegation, and enabling advanced human-agent interactions. 
To address the challenges posed by multi-agent collaboration, we propose the \textbf{Multi-Agent Robotic System (MARS) Challenge}, held at the NeurIPS 2025 Workshop on SpaVLE. The competition focuses on two critical areas: planning and control, where participants explore multi-agent embodied planning using vision-language models (VLMs) to coordinate tasks and policy execution to perform robotic manipulation in dynamic environments. 
By evaluating solutions submitted by participants, the challenge provides valuable insights into the design and coordination of embodied multi-agent systems, contributing to the future development of advanced collaborative AI systems.
\end{abstract}
\section{Introduction}
\label{sec:intro}
Recent progress in multimodal large language models~\cite{team2025gemini,team2025gemma} and vision-language-action models~\cite{black2025pi_,openvla} has driven significant developments in Embodied AI. 
While substantial progress has been made on specific tasks~\cite{cheang2025gr,generalist2025gen0}, the field is now moving towards more complex task scenarios, where multi-agent system frameworks are increasingly necessary.
This shift is driven by three key factors: 
(1) \textbf{Capabilities}. As the capabilities of individual agents increase, scaling them up within a multi-agent framework allows for more complex and versatile systems. 
(2) \textbf{Efficiency}. While individual agents are constrained to specific tasks, multi-agent systems enable the delegation of tasks among agents, making the overall system more efficient. 
(3) \textbf{Human-Agent Interaction}. As embodied multi-agent systems become more capable, they pave the way for advanced human-agent interactions, where embodied systems can collaborate with humans in a broader range of tasks.

To achieve an embodied system, two key capabilities are required: planning and control. Planning is responsible for determining the sequence of actions needed to achieve a particular goal, taking into account the dynamic and often uncertain nature of the environment~\cite{li2025cooperativemultiagentplanningadaptive}. On the other hand, control involves executing these actions by ensuring that each agent moves and interacts with the physically environment in a coordinated and efficient manner~\cite{qin2025robofactory}. Achieving these capabilities in embodied multi-agent systems is even more challenging. Unlike single-agent systems, where the focus is on a solitary entity, multi-agent systems require coordination, communication, and collaboration among agents to complete tasks, introducing additional complexity.

To address these challenges, we proposed the \textbf{Multi-Agent Robotic System (MARS) Challenge},\footnote{\url{https://mars-eai.github.io/MARS-Challenge-Webpage/}} held at the NeurIPS 2025 Workshop on Space in Vision, Language, and Embodied AI.\footnote{\url{https://space-in-vision-language-embodied-ai.github.io/}} 
This competition focuses on two key aspects: planning and control. It aims to advance research on multi-agent collaboration in robotics, where agents of various types, such as humanoids, quadrupeds, and manipulators, must coordinate to achieve complex tasks in dynamic environments. Planning Track explores multi-agent embodied planning, where participants use vision-language models (VLMs) to select agents and define high-level action sequences for collaborative tasks in environments with multiple candidate robots. Control Track focuses on policy execution, requiring participants to deploy end-to-end policy on robotic arms in physically realistic simulations to perform manipulation tasks like multiple blocks stacking, while ensuring robust coordination across agents under partial observability and randomized conditions.

We systematically study multi-agent challenges at both the planning and control levels. Through evaluation and analysis of the participants' submitted solutions, we provide insights that contribute to the development of embodied multi-agent systems, offering new perspectives for advancing the field.
\section{MARS Challenge Overview}
\subsection{Challenge Description}
The Multi-Agent Robotic System (MARS) Challenge is designed to benchmark and advance research on embodied multi-agent collaboration by evaluating both high-level planning and low-level control in heterogeneous robotic systems. Participants are tasked with solving complex multi-agent problems through two complementary tracks: the Planning Track, where models need to select appropriate agents and generate high-level action sequences for collaborative tasks based on visual and language inputs, and the Control Track, where participants need to deploy a control policy on multiple agents to execute manipulation tasks in physically realistic simulation environments. By separating planning and control, the challenge fosters progress in both multi-agent reasoning and coordinated execution, offering insights into the capabilities and limitations of current embodied multi-agent approaches.
\subsection{Planning Track}
\label{subsec:plan_track}

\subsubsection{Task Setup and Evaluation}
The Planning Track of the MARS Challenge\footnote{\url{https://github.com/MARS-EAI/VIKI-R/tree/MARS-Challenge-2025}} targets high-level embodied planning in multi-agent systems with heterogeneous embodiments.
Participants are required to design planners that jointly reason over natural language instructions and visual observations, select appropriate agents, and generate coordinated action sequences.
Unlike low-level control, this track emphasizes semantic task understanding, agent capability reasoning, and long-horizon multi-agent coordination.

Given an instruction and a scene observation, a valid solution must determine both which robots should participate in the task and how they should act over time.
Robot selection contributes 10\% to the final score, while the remaining 90\% evaluates the quality of the generated action plan.
The planning score is computed using a composite metric that considers exact step-wise matching, correctness of the initial action prefix, consistency of action types, and the step length.
This protocol encourages early decision correctness, coherent long-horizon planning, and effective parallel execution.

The Planning Track is built upon a unified embodied intelligence stack. We build on VIKI-Bench~\cite{kang2025viki}, a benchmark for evaluating vision-based embodied planning in multi-agent settings, and instantiate all tasks in the ManiSkill3 physics-based simulator~\cite{tao2024maniskill3}. Task scenarios are drawn from the RoboCasa dataset, which provides diverse household environments and everyday manipulation tasks~\cite{robocasa2024}.
Each task instance consists of a natural language instruction, which may include conditional or exploratory descriptions, and a visual observation depicting the scene and available robots.
The planner outputs a temporally ordered action plan, where one or more robots may act in parallel, and each action is defined by an action type and a target object. More details can be found in Appendix~\ref{sec:appendix_planning}.

The benchmark includes tasks with varying levels of difficulty.
Simple tasks, such as opening an appliance, can often be solved by a single robot with a short action sequence.
In contrast, complex tasks require coordinated execution by multiple robots over long horizons.
A representative example is \texttt{task\_147}, which instructs agents to transport multiple food items into a refrigerator.
The most complex tasks involve planning up to ten steps, highlighting the long-horizon multi-agent tasks.

\subsubsection{Results and Failure Analysis}
\begin{figure*}[t]
    \centering
    \begin{minipage}[t]{0.33\textwidth}
        \vspace{0.3pt}
        \centering
        \scriptsize
        \setlength{\tabcolsep}{10pt}
        \begin{tabular}{clc}
            \toprule
            \textbf{Rank} & \textbf{Team Name} & \textbf{Score}  \\
            \midrule
            1 & EfficientAI & 0.893  \\
            2 & TrustPath AI & 0.858  \\
            3 & jizhiruwo & 0.722 \\
            4 & yeyeyeyeye & 0.703  \\
            5 & jn12xxh & 0.659 \\
            6 & FAITA Group & 0.586  \\
            7 & A & 0.560  \\
            8 & HCB5206 & 0.511  \\
            9 & OpenXS & 0.500  \\
            10 & Sing, dance... & 0.472  \\
            \bottomrule
        \end{tabular}
        \vspace{8pt}
        \captionof{table}{Top 10 Scores}
        \label{tab:score_distribution_plan}
    \end{minipage}%
    \hfill
    \begin{minipage}[t]{0.33\textwidth}
        \vspace{0pt}
        \centering
        \includegraphics[width=\linewidth]{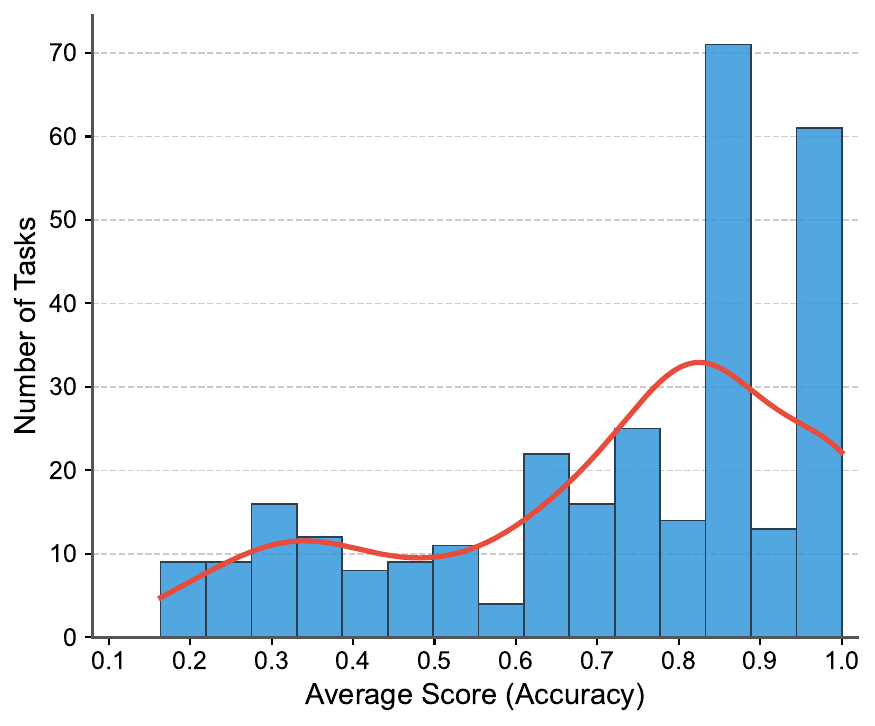}
        \captionof{figure}{Task Accuracy}
        \label{fig:task_accuracy_plan}
    \end{minipage}%
    \hfill
    \begin{minipage}[t]{0.33\textwidth}
        \vspace{0pt}
        \centering
        \includegraphics[width=\linewidth]{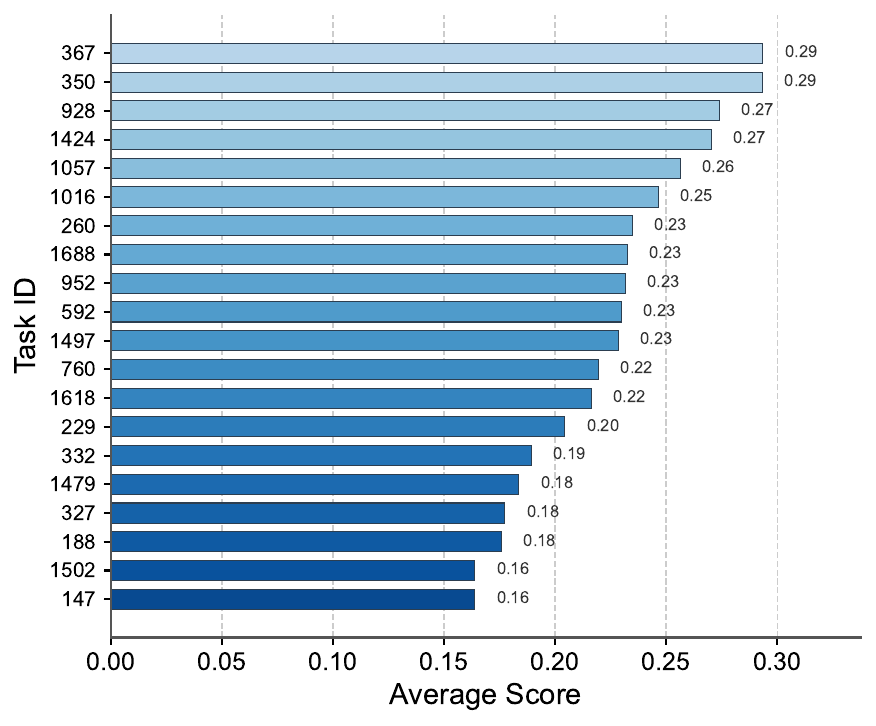}
        \captionof{figure}{Top 20 Hardest Tasks}
        \label{fig:hard_tasks_plan}
    \end{minipage}
    \label{fig:plan_track_results}
    \vspace{-15pt}
\end{figure*}

We analyze participant submissions on the leaderboard to assess planning performance across tasks.
As shown in Tab.~\ref{tab:score_distribution_plan}, overall scores concentrate in the 0.4--0.6 range, indicating substantial headroom under the track’s evaluation criteria.
Fig.~\ref{fig:task_accuracy_plan} provides a more fine-grained view: for many tasks, a large fraction of teams achieve non-zero accuracy, suggesting that generating \emph{feasible} action sequences is often within reach; however, high-accuracy tasks (where most teams solve the instance cleanly) are comparatively sparse, and performance drops markedly on a non-trivial subset of tasks.
Taken together, these patterns imply that the primary bottleneck is not merely producing valid plans, but producing \emph{high-quality} plans that satisfy feasibility while also leveraging multi-agent parallelism and minimizing redundant or unnecessary steps---i.e., efficiency and coordination remain challenging even when task completion is possible.
For completeness, we also identify the 20 hardest tasks in Fig.~\ref{fig:hard_tasks_plan}, highlighting scenarios where even top-performing teams struggle.

Analysis of failure cases reveals several recurring difficulties.
Ambiguous instructions such as ``carry the food'' implicitly require comprehensive scene understanding, where missing even a single relevant object can significantly degrade performance.
Effective solutions demand parallel action assignment across agents, whereas sequential planning underutilizes available robots and leads to lower scores.
Finally, long-horizon tasks amplify the impact of early planning errors due to prefix-based evaluation, making initial agent selection and action ordering particularly critical.
These findings indicate that the Planning Track evaluates not only action prediction accuracy, but also collaborative efficiency and holistic multi-agent reasoning.



\subsection{Control Track}
\subsubsection{Round-wise Rules}



\begin{figure}[h] 
    \centering
    \includegraphics[width=1\textwidth]{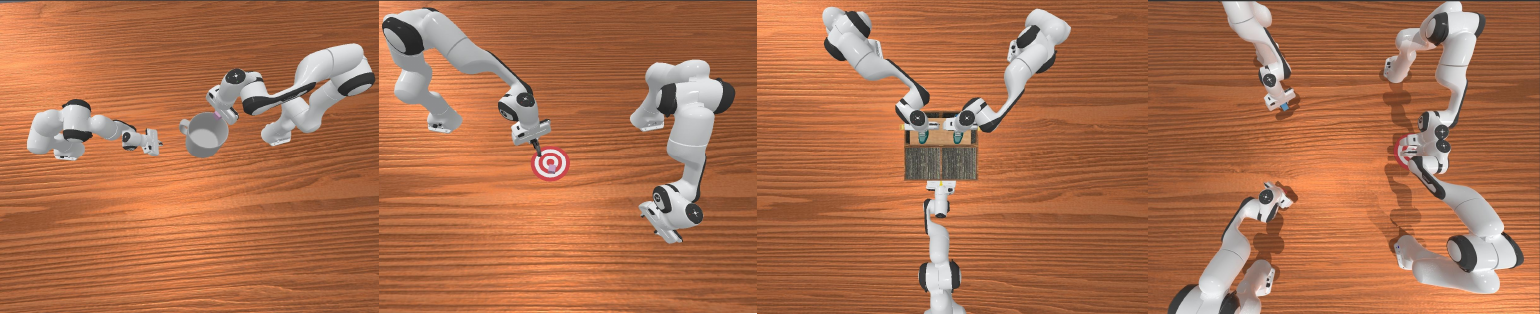} 
    \caption{Visualization of Control Track tasks, including four tasks covering 2, 3, and 4 robotic arms.} 
    \label{fig:mars_control_task}
\end{figure}

The Control Track\footnote{\url{https://github.com/MARS-EAI/RoboFactory/tree/MARS-Challenge-2025}} is dedicated to assessing the collaborative capabilities of multi-agent systems in executing complex tasks within dynamic environments. As shown in Fig.~\ref{fig:mars_control_task}, we have designed four tasks that require robotic arm collaboration: Place Cube in Cup, Strike Cube (Hard), Three Robots Place Shoes, and Four Robots Stack Cube. Each task requires that the agents possess perception and decision-making abilities to interact with the environment and other agents in real-time, under conditions of partial observability and randomization within a volatile setting. In this track, all robots are operated via joint position control, and while each task can be evaluated using a separate set of weights, all solutions must utilize a unified model architecture to ensure cross-task consistency. To support training, we provide an expert data generation pipeline for synthesizing demonstration data, which directly maps observations to actions without the need for a separate high-level planning module. All task data collection is facilitated by RoboFactory~\cite{qin2025robofactory}, built upon ManiSkill~\cite{tao2024maniskill3} and specifically tailored for creating complex, multi-stage tasks that necessitate low-level, collaborative control among multiple robots.


In the testing phase, each task will be tested 100 times, and the average of the success rates for each task will be used as the participant's score. Participants were permitted to collect an unlimited amount of data and were free to set the camera position and data modalities (RGB, depth, point cloud). To ensure fair evaluation, all test-time seeds were kept unseen during the process. Furthermore, the final inference models run on a single RTX 4090 GPU, enforcing a standardized computational constraint.

\subsubsection{Results Overview}

As illustrated in Table \ref{tab:robot_results}, the performance of the top three teams in the control track reveals the inherent limitations of current methodologies. Scores across all tasks fell short of expectations. Some progress was made in dual-arm tasks. However, in the final two tasks, which demand high levels of coordination, nearly all models failed to achieve the specified objectives. Given that these tasks require precise cooperation among three or more robotic arms within a shared workspace, models must contend with exponentially expanding action spaces. These experimental results clearly demonstrate that existing control frameworks still lack sufficient robustness and generalization capabilities when addressing high-dimensional multi-agent collaborative tasks.


\begin{table}[htbp]
\vspace{-2mm}
\centering
\caption{Task success rates (out of 100 trials) of the top three teams in the Control Track.}
\label{tab:robot_results}
\begin{tabular}{lcccc} 
\toprule
\textbf{Task} & \textbf{MMLab@HKUxD-Robotics} & \textbf{INSAIT} & \textbf{RoboStar} & \textbf{Average} \\ 
\midrule
Place Cube in Cup        & 57\% & 13\% & 15\% & 28.3\% \\
Strike Cube (Hard)         & 12\% & 9\% & 7\% & 9.3\% \\
Three Robots Place Shoes & 0\% & 2\% & 1\% & 1.0\% \\
Four Robots Stack Cube   & 0\% & 0\% & 0\% & 0.0\% \\
\bottomrule
\end{tabular}
\end{table}
\section{Example Solutions}
\subsection{Planning Track}
\label{subsec:excellent_plan}

This section highlights two representative solutions from the Planning Track that achieved top performance on the leaderboard.
Both approaches demonstrate strong capabilities in long-horizon reasoning and multi-agent coordination, while adopting different strategies to address the challenges.

\begin{figure}[t]
    \centering
    \includegraphics[width=1\linewidth]{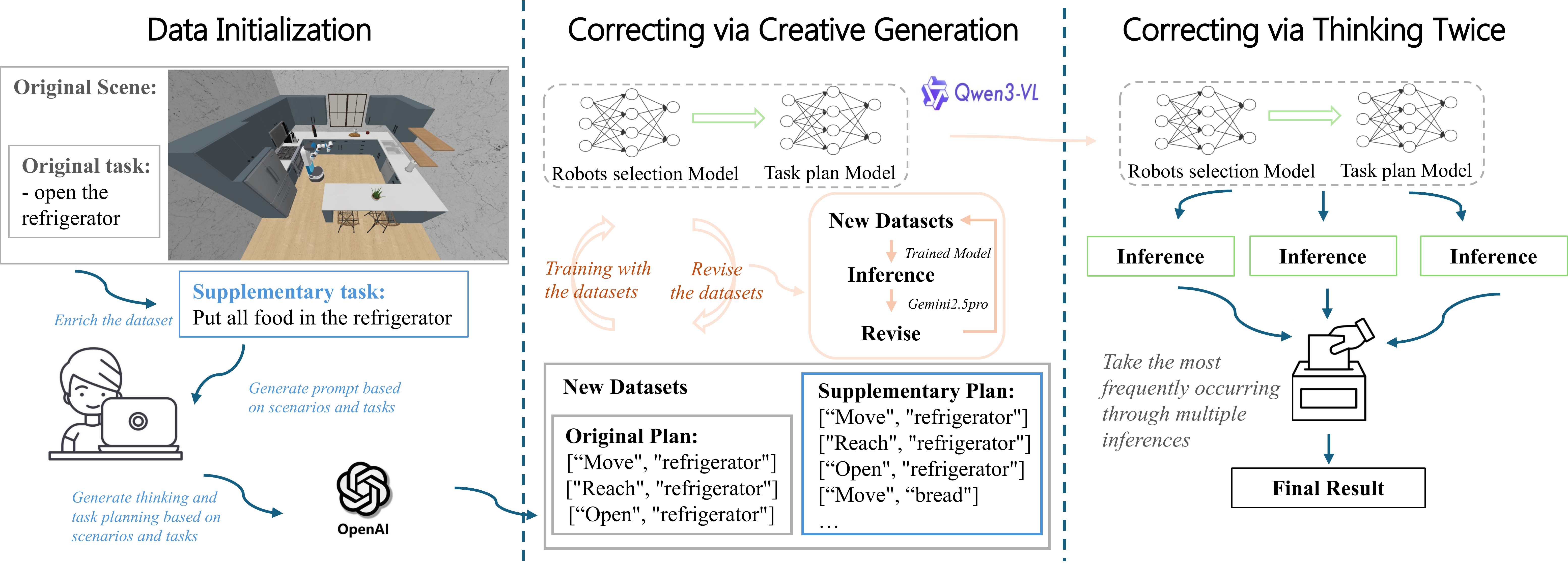}
    \caption{Overall pipeline of the Self-Correction framework. Starting from $N$ manually annotated examples as prompts, VLMs generate seed training data from task instructions and scene observations. Generated plans are evaluated by a judging VLM and refined via supervised fine-tuning using the best plans. This iterative process improves planning performance, while multiple inferences and a voting mechanism address non-unique solutions, resulting in a scalable planning framework.}
    \label{fig:mainarc}
    \vspace{-3mm}
\end{figure}

\paragraph{Scaling Embodied Planning via Self-Correction (Champion).}
The solution introduces a self-correction framework that scales embodied planning ability by explicitly leveraging the creativity and iterative refinement capabilities of vision-language models (VLMs).
As illustrated in Fig.~\ref{fig:mainarc}, instead of relying on a single deterministic planning pass, the method treats planning as an evolving process, where candidate solutions are continuously improved through generation, evaluation, and consensus.

A key insight of this approach is that multi-agent embodied planning often admits multiple valid solutions, and forcing a single inference can trap the model in suboptimal local optima.
To address this, the framework performs multiple stochastic inferences to generate diverse candidate plans and then applies a voting mechanism to select a consensus solution.
This ``thinking twice'' strategy significantly improves robustness in scenarios involving agent heterogeneity and complex assignments.

In addition, the method tackles data scarcity and task heterogeneity through a self-correcting data generation pipeline.
Starting from a small set of manually annotated seed tasks, the planner generates alternative plans, which are subsequently evaluated by a judging VLM.
Higher-quality plans are retained to iteratively refine the planner via supervised fine-tuning.
This closed-loop data expansion process enables the model to progressively adapt to diverse task distributions without requiring large-scale manual annotation.
Overall, the self-correction framework demonstrates that combining creative generation, consensus inference, and iterative data refinement can substantially enhance multi-agent planning performance under ambiguity and non-unique solution spaces.

\begin{figure}[t]
    \centering
    \includegraphics[width=0.98\textwidth]{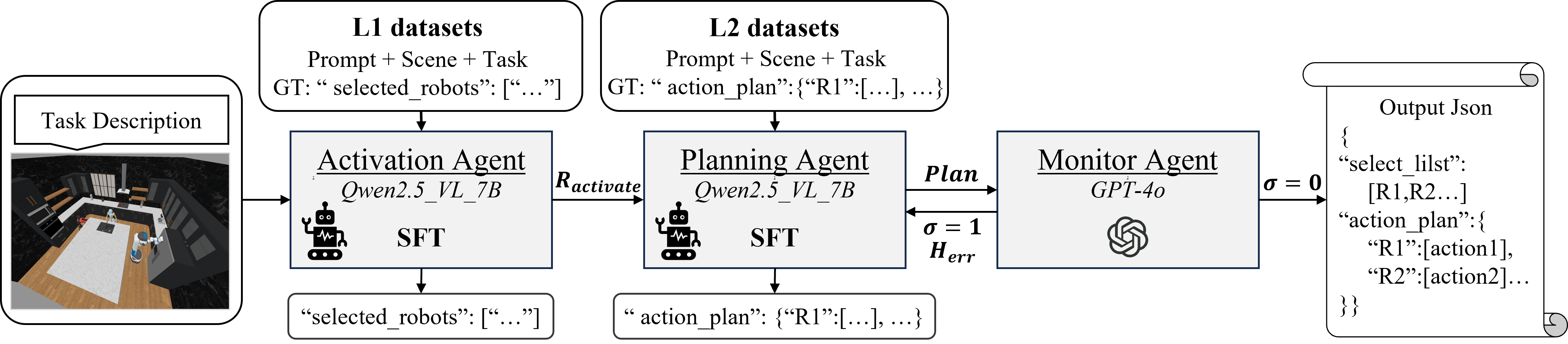}
    \caption{The multi-agent planning system comprises three collaborative components: Activate Agent, Planning Agent and Monitor Agent. Upon receiving a user instruction and a scene image, the system selects the appropriate robots and generates a step-by-step execution plan. Specifically, the Activate Agent and Planning Agent are supervised fine-tuned using the L1 (activation tasks) and L2 (planning tasks) datasets, respectively derived from the adjusted VIKI benchmark~\cite{kang2025viki}.}
    \vspace{-2mm}
    \label{MAS-plan}
\end{figure}

\paragraph{Modular Closed-Loop Framework for Multi-Agent Coordination (Runner-Up).}
The solution adopts a complementary philosophy, focusing on structural decomposition and explicit symbolic grounding. Rather than relying on a monolithic planner, this approach formulates multi-agent planning as a closed-loop system composed of specialized functional modules, each responsible for a distinct reasoning role, enabling more efficient and scalable coordination, as shown in Fig.~\ref{MAS-plan}.

Specifically, the framework separates agent selection, action sequencing, and plan verification into three dedicated, interdependent components.
An activation module first reasons over the task instruction, visual observation, and predefined robot capability priors to select an appropriate subset of agents.
Conditioned on these activated agents, a planning module generates temporally coherent, optimized, and parallelized action sequences under explicit embodiment constraints.
Finally, a monitor module verifies the generated plan for syntactic validity, capability alignment, and logical feasibility, providing corrective feedback when inconsistencies or conflicts are detected.

This modular design mitigates cognitive overload commonly observed in single-model planners and reduces hallucination caused by domain misalignment.
By explicitly encoding robot capabilities and enforcing causal dependencies between actions, the framework ensures that generated plans remain executable within the simulation environment.
Furthermore, the closed-loop interaction between planning and verification enables iterative refinement, improving robustness on long-horizon tasks.
The MAS-Plan framework illustrates the effectiveness of decomposing complex embodied planning into interpretable and verifiable reasoning stages for heterogeneous multi-agent systems.


\subsection{Control Track}
This section presents two solutions from the Control Track. Both approaches demonstrate strong capabilities in language-conditioned multi-arm manipulation and coordination under continuous control, while adopting different strategies to handle combinatorial collaboration in various tasks.

\begin{figure*}[t]
     \centering
     \includegraphics[width=1\linewidth]{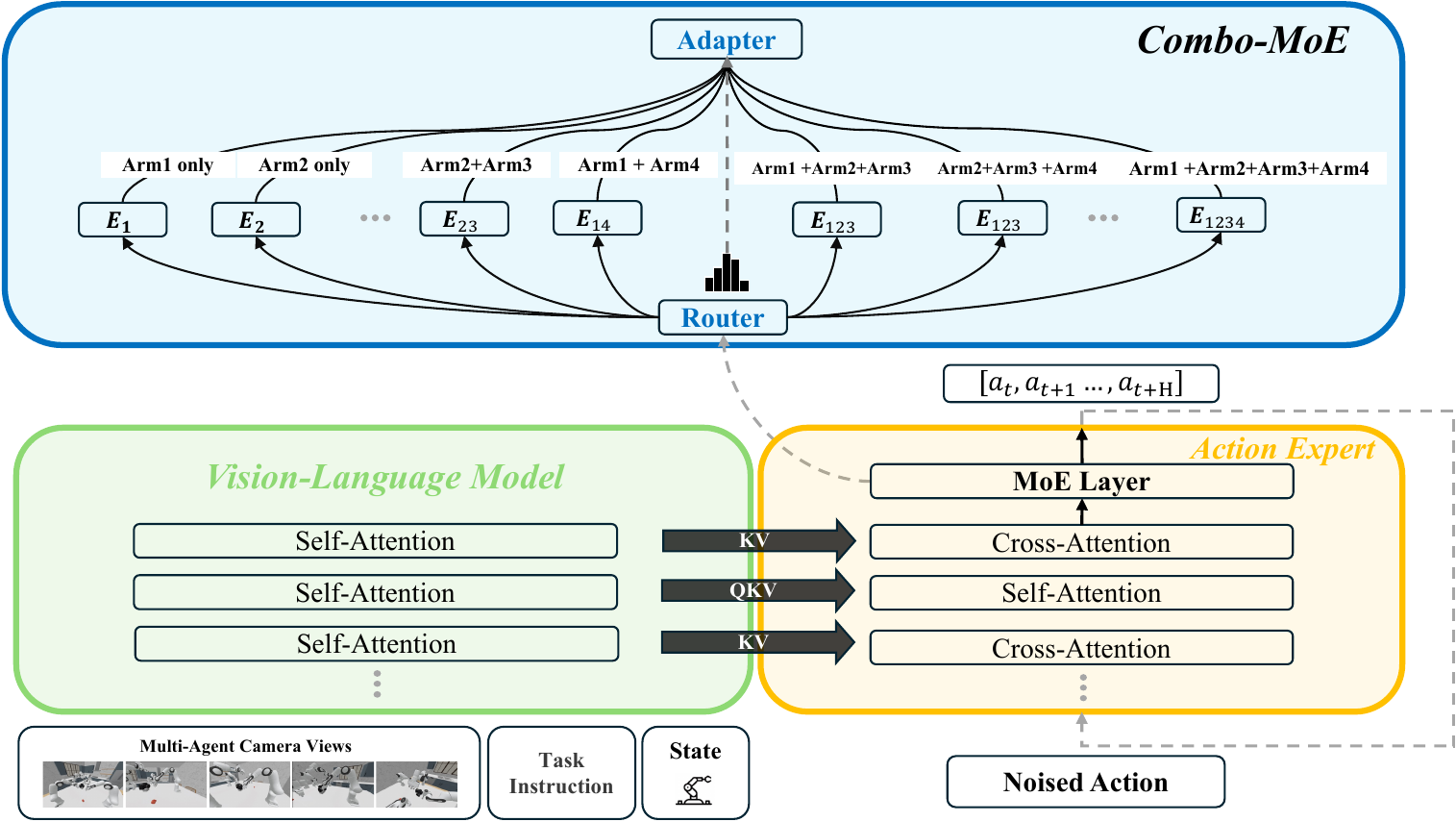}
     \caption{\textbf{Overview of Combo-MoE}. A pre-trained vision-language model encodes the natural-language task instruction and multi-view RGB observations into a unified latent space. On top of this backbone, a mixture-of-experts action head is applied: a router predicts routing weights over subset experts $E_S$ (e.g., single-arm experts $\mathbf{E}_1, \mathbf{E}_2$ and collaboration experts $\mathbf{E}_{23}, \mathbf{E}_{1234}$), an adapter fuses their outputs per arm, and the final action layer decodes an action sequence for all arms.}
     \label{fig:pipeline_combomoe}
     \vspace{-3mm}
\end{figure*}

\paragraph{Combo-MoE: Combinatorial Experts for Multi-Arm Coordination (Champion)}
The winner solution is Combo-MoE, a combinatorial mixture-of-experts (MoE) architecture that enables scalable multi-arm collaborative manipulation. The solution is built on the principle that multi-arm coordination is inherently discrete and combinatorial. As demonstrated in Fig.~\ref{fig:pipeline_combomoe}, Combo-MoE factorizes the multi-arm action space by instantiating subset-specific experts for all non-empty arm combinations, resulting in $2^N - 1$ experts for $N$ arms. Each expert specializes in behaviors where a particular subset of arms is active, ranging from single-arm primitive skills to higher-order coordination patterns such as handover, shared-object manipulation, and collision-aware cooperation. This structured decomposition preserves modularity to avoid manual task decomposition.

The policy is built on top of a pre-trained vision-language model (VLM), which encodes language instructions (e.g., “grab the steak and use the camera to photograph it”) and multi-view RGB observations into a unified latent space. The shared backbone for all experts ensures consistent reasoning over semantically grounded representations, enabling effective skill reuse across tasks and arm configurations. Subsequently, a modular router–adapter MoE layer is applied as an action head. The router predicts non-exclusive activation weights over subset experts, allowing multiple experts (e.g., single-arm and multi-arm coordination experts) to be simultaneously active. The adapter then performs per-arm action fusion, selectively combining expert outputs to produce coherent multi-arm action sequences. This design avoids naive averaging and allows the policy to dynamically balance independent control and tight collaboration based on task context.

To align learning with the combinatorial architecture, the solution adopts a three-stage training strategy. Firstly, it performs individual expert pretraining, where each subset expert is trained via behavior cloning on demonstrations filtered by the active arms. Secondly, it proceeds to router–adapter learning, where all experts are frozen and the MoE layer is fine-tuned to select and compose experts for full multi-arm behavior manipulation. Finally, it applies joint finetuning on the experts and MoE layer, enabling adaptations between collaborative tasks while preserving expert specialization.

Combo-MoE provides a scalable, modular, and interpretable solution for collaborative manipulation. By making the policy explicit and learnable, this approach significantly reduces learning complexity and offers a promising path toward scaling robotic manipulation in complex multi-arm systems.

\begin{figure*}[t]
     \centering
     \includegraphics[width=1\linewidth]{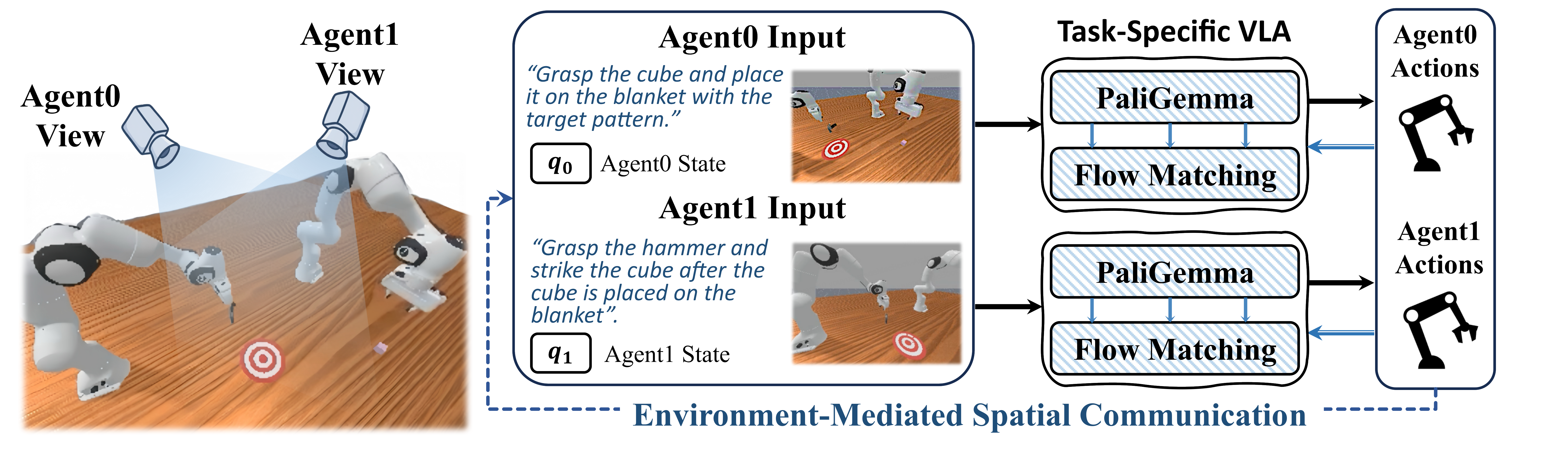}
          \caption{\textbf{Overview of CoVLA.} CoVLA formulates multi-agent manipulation as a decentralized problem. Each robot arm executes an independent policy, with coordination achieved through visual perception of the shared manipulation workspace.}
     \label{fig:pipeline_CoVLA}
     \vspace{-5mm}
\end{figure*}

\paragraph{CoVLA: Collaborative VLA via Decentralized Manipulation (Runner-Up)}
The runner-up solution proposes a Collaborative Vision-Language-Action (CoVLA) for multi-robot manipulation. As shown in Fig.~\ref{fig:pipeline_CoVLA}, this framework treats coordination as a decentralized, game-theoretic information-sharing and reward-shaping problem. Instead of learning a centralized joint policy, CoVLA decomposes collaboration across independently controlled robots, each operating its own task-specialized VLA policy. This design directly addresses the challenges of semantic task understanding and coordination in complex, language-conditioned multi-robot manipulation scenarios.

CoVLA is built on top of the $\pi_{0}$~\cite{black2024pi_0} foundation model, where each robot arm runs an independent fine-tuned instance for its specific sub-task. Leveraging $\pi_{0}$'s pre-trained backbone, each agent inherits strong semantic, spatial, and temporal reasoning capabilities, enabling it to interpret complex instructions such as multi-stage or conditional task descriptions. Compared to diffusion-based policies that rely on low-level state representations, CoVLA provides richer semantic grounding at the cost of reduced joint controllability, which is mitigated through task-specific fine-tuning.

Coordination in CoVLA is achieved through environment-mediated spatial communication, rather than explicit inter-agent messaging. Each agent observes the shared workspace through its own third-person camera view, allowing changes in the environment caused by one robot’s actions to be implicitly perceived by others. This shared visual grounding enables robust coordination across space and time, such as synchronizing object placement and striking actions or avoiding collisions during collaborative manipulation, while preserving decentralized execution.

Each task-specific VLA is trained independently using role-specific demonstrations collected from the agent’s own camera viewpoint. The reward shaping within each agent’s training data encourages both correct task sequencing (e.g., verifying conditions before acting) and precise manipulation, leveraging privileged information during training to improve robustness at deployment. This distributed framework allows each agent to specialize in its assigned manipulation role while maintaining alignment with the global task through shared visual context.

Overall, CoVLA offers a decentralized approach to multi-robot collaboration, combining pre-trained vision-language reasoning with environment-mediated coordination. By replacing centralized joint action modeling and with independent VLAs with shared visual grounding, CoVLA achieves flexible, scalable collaboration for language-conditioned multi-robot manipulation tasks. This makes it a strong and complementary alternative to centralized combinatorial policy architectures.
\section{Discussion}
\subsection{Key Findings}
\begin{tcolorbox}[takeawaysbox]
\begin{enumerate}[leftmargin=1em]
\item \textbf{Iterative planning helps because it turns planning errors into recoverable edits:}
re-sampling and revising plans reduces long-horizon compounding mistakes, improving success on tasks with ambiguity or late-stage constraints.
\item \textbf{Multi-agent performance improves either by (i) stronger per-agent skills or (ii) better coordination structure:}
domain-specialized policies improve each agent’s execution reliability, while structured coordination (e.g., expert decomposition / shared grounding) reduces inter-agent conflicts and scales to more arms.
\end{enumerate}
\end{tcolorbox}
\paragraph{What to take away.}
Across top solutions, improvements come from two recurring mechanisms.
(1) \emph{Plan-level self-correction:} iterative planning/revision generates multiple candidate plans, checks feasibility, and revises after observing failures, which reduces error accumulation over long horizons.
(2) \emph{Agent-level reliability and coordination:} domain-specific training increases per-agent success on its sub-skill, while coordination structures (shared grounding, expert decomposition, or explicit role partitioning) reduce conflicts and make scaling to more arms tractable.

Iterative planning improves multi-agent performance primarily by reducing \emph{long-horizon compounding errors} and \emph{cross-agent inconsistency}.
In multi-agent manipulation, an early high-level mistake (wrong subtask order, infeasible handover, or conflicting resource use) often propagates into downstream execution failures because agents commit to incompatible trajectories.
Iterative planning mitigates this by (i) generating multiple candidate plans, (ii) checking them against feasibility signals from the scene (e.g., collisions, reachability, or prerequisite satisfaction), and (iii) revising the plan after observing partial outcomes.
As a result, agents can recover from early misallocations and progressively converge to plans that are both executable and mutually consistent, which is especially beneficial in tasks with ambiguous object states or late-stage constraints.

Two design patterns repeatedly appear in scalable multi-arm systems, especially in the control track.
\emph{Structured coordination} (e.g., combinatorial experts or explicit coordination patterns) improves performance by narrowing the search/control space: it decomposes coordination into reusable interaction templates, reducing the burden of learning all pairwise couplings from scratch and enabling better scaling as the number of arms increases.
\emph{Decentralized execution with shared grounding} improves performance by increasing robustness: each agent runs a task-specialized policy, and coordination emerges through observing the same scene state, which reduces single-point-of-failure planning and allows agents to adapt online when other agents deviate.
In practice, strong systems often combine both: a lightweight coordination structure to prevent conflicts, and decentralized policies to maintain flexibility under partial observability and stochastic contacts.

To avoid method-specific jargon, we summarize these approaches by the underlying mechanism (self-correction vs.\ coordination/reliability), rather than by the names used in individual submissions.

\subsection{Spatial Reasoning and Communication Mechanisms}
In both the planning and control tracks, spatial reasoning and communication play critical roles in multi-agent collaboration. 
For planning, spatial reasoning is primarily achieved through the use of vision-language models (VLMs), which process visual inputs such as multi-view RGB images to understand the positions of agents and objects. Participants typically used VLMs as perceptual backbones to encode multi-view RGB observations into semantic/spatial latent representations (e.g., object states, relative poses, and agent configurations), which were then consumed by downstream planners or policies to produce executable action sequences; in some cases, the VLM was additionally prompted to output high-level textual plans that were subsequently grounded and converted into actions. This allows for coordinated task execution, where agents can plan and adjust actions based on spatial context.
In the control track, the emphasis shifts from reasoning to communication of spatial information. Here, agents rely on local perceptual data to adjust their actions based on changes observed in the shared workspace. Without explicit inter-agent messaging, coordination must occur through implicit environmental cues to avoid collisions and ensure smooth collaboration. This decentralized approach enhances scalability and robustness, as each agent must coordinate based on the evolving spatial context. The MARS Challenge highlights the importance of both spatial reasoning for planning and information exchange for control, advancing the development of multi-agent systems that can effectively collaborate in dynamic environments.

\subsection{Challenges Faced}
At the planning level, one of the major challenges is selecting the optimal strategy for long-horizon tasks. VLMs, while powerful, often struggle with the complexity of these tasks, as they tend to provide suboptimal solutions or "lazy" plans (details on \ref{supp:lazy_plan}). Even with reinforcement learning rewards in place, finding the most effective decision plan is difficult. The challenge lies in ensuring that the system can select and adapt to long-term strategies, where early decisions can significantly affect the overall outcome, and subtle adjustments are needed to optimize performance over time.

At the control level, the increasing number of agents introduces a significant challenge in managing the growing action space and coordinating actions. As the number of agents increases, the action dimensions grow exponentially, making it difficult to maintain effective coordination. Decentralized control, while offering greater flexibility, faces the difficulty of effectively integrating and coordinating the information from multiple agents into a unified policy. This requires innovative solutions to ensure that each agent can collaborate efficiently without causing conflicts or losing track of the overall task, making the coordination of multi-agent systems a key obstacle.
\subsection{Limitations and Future Work}
One of the primary limitations of the current work is that it primarily focuses on simulation environments, which may not fully capture the complexities of real-world scenarios. While simulations provide a controlled setting to test algorithms and strategies, they often overlook factors like noise, physical variability, and unexpected interactions that can arise in real environments. There is a notable gap between simulated performance and real-world applicability, especially in tasks involving physical manipulation and dynamic coordination across multiple agents. Future work should focus on transferring the models and strategies to real-world settings, where they can be validated and further refined. This includes addressing issues related to perception, actuation, and real-time decision-making, which are often more challenging in physical environments.

\section{Related Work}

\paragraph{Embodied Planning in MAS}
The integration of Vision-Language Models (VLMs) into Multi-Agent Systems (MAS) has shifted embodied planning from rigid symbolic logic to semantic, open-world reasoning. VLM-based planners can interpret natural language instructions, decompose tasks, and reason about physical constraints. Centralized frameworks, such as COHERENT~\cite{liu2025coherentcollaborationheterogeneousmultirobot}, decompose long-horizon tasks and optimize allocation based on heterogeneous agent capabilities, while decentralized approaches like RoCo~\cite{mandi2023rocodialecticmultirobotcollaboration} and MADRA~\cite{wang2025madramultiagentdebateriskaware} use VLMs for inter-agent dialogue and consensus-based strategy development. Several studies combine LLMs with structured planning methods such as dependency graphs~\cite{wang2024dart}, linear programming~\cite{obata2024lip}, and actor-critic frameworks~\cite{wang2025multi}. For embodied agents, Co-ELA~\cite{zhang2023building} integrates perception, planning, and execution components, while COMPASS~\cite{li2025cooperativemultiagentplanningadaptive} utilizes visual verification and structured communication to handle partial observability and execution failure. At the same time, embodied multi-agent benchmarks for collaborative evaluation have also evolved. In 2D, LLM-Co~\cite{agashe2025llm} and Overcooked~\cite{carroll2019utility} focus on strategic coordination, while 3D benchmarks like WAH~\cite{puig2020watch} and PARTNR~\cite{chang2024partnr} address social intelligence and collaborative visual planning. Multi-agent manipulation is evaluated by RocoBench~\cite{roco} and FurnMove~\cite{jain2020cordial}. Broader frameworks such as LLaMAR~\cite{nayak2025llamarlonghorizonplanningmultiagent} and VIKI-Bench~\cite{kang2025viki}, which bridge planning and manipulation tasks, set new standards for evaluation. Building on these, the MARS Challenge leverages VIKI-Bench to provide embodied multi-agent planning tasks, advancing the evaluation and development of planning strategies in multi-agent environments.

\paragraph{Robotic Manipulation in MAS}
Specialized policy architectures~\cite{chi2023diffusion,ke20243d,adaptdiffuser,liang2024skilldiffuser,dexhanddiff,wang2024rise,wen2025dexvla,Ze2024DP3} often excel on narrowly defined tasks yet struggle to carry over to new robot embodiments and new environments. In contrast, foundation models trained on million-scale, multi-robot corpora exhibit strong zero-shot transfer: RT-1~\cite{brohan2022rt-1} unifies vision, language, and action in a single transformer for real-time kitchen manipulation; RT-2~\cite{brohan2023rt-2} jointly finetunes large vision–language models on web and robot data to support semantic planning and object reasoning; diffusion-based RDT-1B~\cite{liu2024rdt1b} and $\pi_0$\cite{black2024pi_0} learn diverse bimanual dynamics from over a million episodes. Vision–language–action systems such as OpenVLA\cite{openvla} and CogACT~\cite{li2024cogact}, together with adaptations like Octo~\cite{octo_2023}, LAPA~\cite{lapa}, and OpenVLA-OFT~\cite{openvla_oft}, further demonstrate efficient finetuning across robots and sensing modalities. Collectively, these results point to a data-driven bottleneck: robust cross-task generalization hinges on large, diverse, and high-fidelity datasets that faithfully capture real-world appearance, sensing, and physics. Beyond predominantly single-agent paradigms, recent efforts have begun to investigate multi-agent robotic manipulation. Systems such as RoboBallet~\cite{doi:10.1126/scirobotics.ads1204}, RoboFactory~\cite{qin2025robofactory}, RoboTwin~\cite{chen2025robotwin}, and RoboOS~\cite{tan2025roboos} explore scalable infrastructures for data collection and policy learning in multi-arm settings, emphasizing coordination, synchronization, and collaborative execution. Notably,  RoboBallet~\cite{doi:10.1126/scirobotics.ads1204} proposes a graph neural network–based reinforcement learning framework for joint task allocation, scheduling, and collision-free motion planning of multiple robotic arms in obstacle-rich shared workspaces.
RoboFactory~\cite{qin2025robofactory} introduces a structured benchmark for multi-agent cooperation, enabling systematic evaluation of collaborative manipulation across various task compositions and robot configurations. Building on these efforts, the MARS Challenge provides a benchmark of embodied multi-agent manipulation on common tasks, further fostering evaluation and advancing the development of coordinated multi-agent robotic systems.

\section{Conclusion}
In this work, we explored the challenges and advancements in multi-agent systems through the MARS Challenge at NeurIPS 2025. We focused on two key areas: planning and control, investigating how vision-language models can be used to coordinate multi-agent tasks and how policy execution can be optimized for robotic manipulation in dynamic environments. Our findings highlight the importance of iterative optimization and agent specialization in improving multi-agent collaboration. This work contributes to the ongoing development of scalable, flexible, and efficient multi-agent systems, providing insights for future research in real-world applications and advanced collaboration.
\section{Participants and Committees}
\label{app:participants}

\subsection*{Organizers}
Li Kang, Heng Zhou, Xiufeng Song, Rui Li, Bruno N.Y. Chen, Ziye Wang, Ximeng Meng.

\subsection*{Advisory Committee}
Zhenfei Yin, Stone Tao, Ziqiao Ma, Yiran Qin, Xiaohong Liu, Ruimao Zhang, Lei Bai, Yilun Du, Hao Su, Philip Torr.

\subsection*{Planning Track}

\paragraph{Champion: EfficientAI}
Ruihao Gong$^{\dagger}$, Yejun Zeng$^{\dagger}$, Fengjun Zhong$^{\dagger}$, Shenghao Jin, Jinyang Guo, Xianglong Liu (Beihang University, {\small $^{\dagger}$ Equal contribution.}).

\paragraph{Runner-up: TrustPath AI}
Xiaojun Jia (Nanyang Technological University), Tianqi Shan (Sun Yat-sen University), Wenqi Ren (Sun Yat-sen University), Simeng Qin (Northeastern University), Jialing Yang (Sun Yat-sen University), Xiaoyu Ma (Nanyang Technological University).

\subsection*{Control Track}

\paragraph{Champion: MMLab@HKU $\times$ D-Robotics}
Tianxing Chen (The University of Hong Kong), Zixuan Li (Shenzhen University), Zijian Cai (Shenzhen University), Yan Qin (Hong Kong University of Science and Technology (Guangzhou)), Yusen Qin (D-Robotics), Qiangyu Chen (The City University of Hong Kong), Kaixuan Wang (The University of Hong Kong), Zhaoming Han (D-Robotics), Yao Mu (Shanghai Jiao Tong University), Ping Luo (The University of Hong Kong).

\paragraph{Runner-up: INSAIT}
Yuanqi Yao (INSAIT, Sofia University ``St.\ Kliment Ohridski''), Haoming Song (Shanghai Jiao Tong University), Jan-Nico Zaech (INSAIT, Sofia University ``St.\ Kliment Ohridski''), Fabien Despinoy (Toyota Motor Europe), Danda Pani Paudel (INSAIT, Sofia University ``St.\ Kliment Ohridski''), Luc Van Gool (INSAIT, Sofia University ``St.\ Kliment Ohridski'').

\section*{Acknowledgment}
This challenge is supported by Hillbot Inc. We also thank the organizing committee of the NeurIPS 2025 Workshop on Space in Vision, Language, and Embodied AI (SpaVLE) for helpful discussions and logistical support. We also thank the Shanghai Artificial Intelligence Laboratory for its support.

\appendix

\section*{Appendix}
\section{Planning Track Details}
\label{sec:appendix_planning}

To provide a comprehensive understanding of the Planning Track, we detail the problem formulation, the specific output format, and representative case studies of generated plans.

\subsection{Problem Formulation and Output Format}

The planning problem is defined as generating a solution tuple $(S, P)$ given an instruction $I$ and visual observation $V$, where $S$ is the set of selected robots and $P$ is the temporal action plan.

\subsubsection{Action Space}
An atomic action in the plan is defined as a tuple:
\begin{equation}
    a_{t}^{(r)} = (\text{ActionType}, \text{TargetObject})
\end{equation}
where $r$ denotes the specific robot and $t$ denotes the time step.
\begin{itemize}
    \item \textbf{ActionType}: A high-level primitive operation (e.g., \texttt{Move}, \texttt{Pick}, \texttt{Open}, \texttt{Place}).
    \item \textbf{TargetObject}: The specific instance name of the object to be manipulated (e.g., \texttt{pear}, \texttt{oven}, \texttt{cabinet}).
\end{itemize}

\subsubsection{JSON Submission Structure}
Participants are required to submit solutions in a structured JSON format. This format must strictly adhere to the schema required for automated CSV conversion and scoring. A representative example is shown in Listing~\ref{lst:json_format}.

\begin{lstlisting}[
    language=Caml, 
    caption={Example JSON output for a multi-agent task.}, 
    label={lst:json_format}, 
    basicstyle=\ttfamily\footnotesize, 
    frame=single,
    columns=fullflexible,
    breaklines=true
]
{
  "task_id": "task_1044",
  "answer": {
    "selected_robots": ["Fetch", "Stompy"],
    "action_plan": [
      {
        "step": 1,
        "actions": {
          "Fetch": ["Move", "pear"],
          "Stompy": ["Move", "knife"]
        }
      },
      {
        "step": 2,
        "actions": {
          "Fetch": ["Pick", "pear"],
          "Stompy": ["Pick", "knife"]
        }
      }
    ]
  }
}
\end{lstlisting}

\subsection{Case Studies: Simple vs. Complex Tasks}

The benchmark evaluates planners across a spectrum of difficulty. We contrast a simple single-agent task with a complex multi-agent coordination scenario.

\subsubsection{Simple Task: Sequential Manipulation}
\textbf{Task ID:} \texttt{task\_7} \\
\textbf{Instruction:} ``Carry out the operation of opening the oven.'' \\
\textbf{Selected Robots:} \texttt{['Stompy']}

This task represents a standard short-horizon manipulation problem solvable by a single agent sequentially. The optimal plan is concise, as shown in Table~\ref{tab:task7}.

\begin{table}[h]
    \centering
    \caption{Plan for Task 7 (Simple)}
    \label{tab:task7}
    \begin{tabular}{c c l l}
        \toprule
        \textbf{Step} & \textbf{Agent} & \textbf{Action Type} & \textbf{Target Object} \\
        \midrule
        1 & Stompy & Move & oven \\
        2 & Stompy & Reach & oven \\
        3 & Stompy & Open & oven \\
        \bottomrule
    \end{tabular}
\end{table}

\subsubsection{Complex Task: Collaborative Long-Horizon Transport}
\textbf{Task ID:} \texttt{task\_147} \\
\textbf{Instruction:} ``Carry the food and put it to the refrigerator.'' \\
\textbf{Selected Robots:} \texttt{['Fetch', 'Stompy']}

This task is classified as ``Hard'' with an average score of only $\sim 0.16$ on the leaderboard. It requires highly synchronized parallel execution over a horizon of approximately 10 steps.

\textbf{Key Challenges:}
\begin{enumerate}
    \item \textbf{Ambiguity and Grounding:} The term ``food'' implies a set of objects. The planner must visually identify all relevant items (e.g., \textit{pear, bread, tomato, pumpkin, apple}) from the scene observation. Missing any item results in an incomplete plan.
    \item \textbf{Parallel Coordination:} To achieve a high score, the planner must assign actions to both \texttt{Fetch} and \texttt{Stompy} simultaneously. Sequential plans (one robot waiting while the other acts) are penalized by the \textit{Length Ratio} metric.
    \item \textbf{Prefix Sensitivity:} The scoring metric assigns 30\% weight to \textit{Prefix Match}. An error in the early steps of a 10-step plan significantly degrades the final score.
\end{enumerate}

\subsection{Lazy Plan}
\label{supp:lazy_plan}
We observe that models often output a "lazy plan". Here, a \emph{lazy plan} refers to a low-effort strategy that avoids task-specific reasoning and step minimization: to maximize the chance of success, the planner executes a broad set of redundant actions (often covering all plausible options) rather than inferring the most likely state and committing to an efficient sequence.
For example, consider a task that asks the robot to \emph{place the apple on the table into the bowl}. A lazy plan may over-generalize the instruction and execute a “cover-all” routine: it starts picking and placing \emph{every} visible fruit (e.g., apples, pears) into the bowl, and may even waste steps searching for or attempting to manipulate objects that are not present, instead of first identifying the target apple and executing the minimal pick$\rightarrow$place sequence. In this case, the model does not reason about \emph{what is necessary vs.\ unnecessary} for task completion, and trades deliberate selection for redundant actions.

\medskip

{
\small
\bibliographystyle{plain}
\bibliography{main}



}





\newpage

\end{document}